\documentclass[runningheads]{llncs}
\usepackage[T1]{fontenc}
\usepackage[utf8]{inputenc}
\usepackage{amsmath}
\usepackage[colorlinks=true, linkcolor=blue, citecolor=blue, urlcolor=blue]{hyperref}

\usepackage{placeins}

\usepackage{graphicx}
\begin{document}

\thispagestyle{empty} 

\title{Evaluating OCR performance on food packaging labels in South Africa}

\titlerunning{Evaluating OCR performance on food packaging labels in South Africa}

%

\author{Mayimunah Nagayi\inst{1}\orcidID{0009-0009-9241-3059} \and Alice Khan \inst{2} \orcidID{0000-0003-1011-3693} \and Tamryn Frank \inst{2} \orcidID{0000-0002-5180-9171} \and Rina Swart \inst{3} \orcidID{0000-0002-7786-3117} \and Clement Nyirenda\inst{1,4}\orcidID{0000-0002-4181-0478}} 
\authorrunning{M Nagayi et al.}
%

\institute{
Department of Computer Science, University of the Western Cape, Robert Sobukwe Rd, Bellville, Cape Town, 7535, South Africa, \email{4163113@myuwc.ac.za}
\vspace{0.5em}
\and
School of Public Health, University of the Western Cape, Robert Sobukwe Rd, Bellville, Cape Town, 7535, South Africa, \email{askhan@uwc.ac.za}, \email{tfrank@uwc.ac.za}
\vspace{0.5em}
\and
Department of Dietetics and Nutrition, University of the Western Cape, Robert Sobukwe Rd, Bellville, Cape Town, 7535, South Africa, \email{rswart@uwc.ac.za}
\vspace{0.5em}
\and
eResearch Office, University of the Western Cape, Robert Sobukwe Rd, Bellville, Cape Town, 7535, South Africa, \email{cnyirenda@uwc.ac.za}
}
\maketitle              

\begin{abstract}
This study evaluates four open-source Optical Character Recognition (OCR) systems which are Tesseract, EasyOCR, PaddleOCR, and TrOCR on real world food packaging images. The aim is to assess their ability to extract ingredient lists and nutrition facts panels. Accurate OCR for packaging is important for compliance and nutrition monitoring but is challenging due to multilingual text, dense layouts, varied fonts, glare, and curved surfaces. A dataset of 231 products (1{,}628 images) was processed by all four models to assess speed and coverage, and a ground truth subset of 113 images (60 products) was created for accuracy evaluation. Metrics include Character Error Rate (CER), Word Error Rate (WER), BLEU, ROUGE-L, F1, coverage, and execution time. On the ground truth subset, Tesseract achieved the lowest CER (0.912) and the highest BLEU (0.245). EasyOCR provided a good balance between accuracy and multilingual support. PaddleOCR achieved near complete coverage but was slower because it ran on CPU only due to GPU incompatibility, and TrOCR produced the weakest results despite GPU acceleration. These results provide a packaging-specific benchmark, establish a baseline, and highlight directions for layout-aware methods and text localization.

\keywords{OCR \and Food Packaging \and Ingredient Lists \and Nutrition Facts Panels \and Tesseract \and EasyOCR \and PaddleOCR \and TrOCR.}

\end{abstract}

\section{Introduction}

Consumers rely on printed information on food packaging such as ingredient lists, allergen warnings, and nutrition facts to make informed dietary decisions. While regulations require this information to be accurate and legible, packaging often reduces readability with cluttered layouts, small fonts, decorative text styles, and multilingual content for export. These design choices make it harder for people to read and create additional challenges for digital processing. Child-directed marketing practices, which often include illustrations and fantasy imagery, further reduce legibility, particularly in breakfast cereals and snacks \cite{khan2023}. For example, multilingual text and inconsistent formatting complicate both manual review and automated extraction, limiting the accuracy of nutrition apps, compliance checks, and large-scale health research \cite{guimaraes2023,seitaj2024}. This context highlights the need for Optical Character Recognition (OCR) systems,  which convert text in images into machine readable text, that can handle real world packaging conditions rather than controlled document layouts \cite{abdoolkarim2025}.

Accurate extraction of ingredient lists and nutrition facts panel text from retail food packaging remains difficult under real world capture conditions. The target application is large scale extraction from shelf images and product photos to support consumer transparency and regulatory checks in South Africa, benefiting regulators, retailers, public health researchers, and consumer tools. The objective is to compare four open source OCR systems on these tasks and to report accuracy at character and word levels, product level coverage, and processing time per image. Real world images add complexity through glare, shadows, and curved or reflective surfaces that distort text \cite{rodin2019}. Design and imaging factors make packaging more complex than standard documents \cite{guimaraes2023}. Seitaj and Elangovan \cite{seitaj2024} further note that mixed text styles and multilingual content frequently cause segmentation and recognition errors. Many OCR models struggle when text switches language within the same block \cite{huang2021}.

Although OCR systems perform well on scanned documents and scene-text benchmarks, their reliability on food packaging remains underexplored. Public datasets and competitions such as the International Conference on Document Analysis and Recognition (ICDAR) focus on either documents or natural scenes, which lack the dense tabular structures and multilingual content typical of packaging \cite{karatzas2013}. Large-scale benchmarks like TextOCR emphasize arbitrarily shaped text in outdoor images but do not include nutrition panels or ingredient lists \cite{singh2021}. Recent advances such as multilingual OCR networks and layout-aware frameworks \cite{huang2021,shu2024} aim to improve recognition across scripts and structured regions, but these methods are rarely tested on packaging with language switching and fine-grained text. Existing packaging studies, including those by Guimarães et al. \cite{guimaraes2023} and Rosyadi et al. \cite{rosyadi2024}, highlight the complexity of cluttered layouts and mixed fonts but rely on small datasets or single-engine evaluations. These limitations make it difficult to establish fair comparisons or draw conclusions about real world performance, creating a clear need for systematic benchmarking under uncontrolled imaging conditions.

This paper addresses these gaps by presenting a comparative evaluation of four widely used open-source OCR systems: Tesseract, EasyOCR, PaddleOCR, and TrOCR on real world food packaging images captured in South African retail environments. The study focuses on ingredient lists and nutrition facts panels using full packaging images and applies a standardized post-processing pipeline for normalization. Performance is assessed using character-level, word-level, and semantic metrics alongside coverage and execution time, establishing a domain-specific benchmark for packaging OCR.

This paper advances packaging OCR research in several ways. First, it evaluates Tesseract, EasyOCR, PaddleOCR, and TrOCR on real world food packaging, focusing on ingredient lists and nutrition facts panels. Second, it introduces a benchmark that uses multiple metrics, including CER, WER, BLEU, ROUGE‑L, F1, coverage, and execution time. Third, it applies a standardized post‑processing workflow to normalize outputs across models, and finally, it offers insights into performance trade offs and limitations under practical deployment conditions.

The remainder of the paper is organized as follows: Section 2 provides an overview of the OCR systems evaluated in this study. Section 3 presents the methodology, including dataset description, preprocessing, and experimental setup. Section 4 outlines the evaluation metrics and implementation details, while Section 5 reports and discusses the results, and finally, Section 6 concludes the paper and highlights potential directions for future research.

\section{OCR Approaches}
\subsection{Tesseract OCR}
Tesseract is one of the most widely used open-source OCR engines, initially developed by Hewlett-Packard and later maintained by Google. Early versions relied on rule-based segmentation, which restricted flexibility for non-standard layouts. Since version 4.0, Tesseract incorporates an LSTM-based recognition module, allowing sequence learning and improving accuracy on printed text compared to its earlier design \cite{smith2007}. This architecture is effective on clean, structured text, making Tesseract a strong baseline for document-oriented OCR tasks.

Despite these advances, Tesseract remains highly sensitive to noisy backgrounds, small fonts, and curved text regions. Its reliance on line segmentation causes recognition errors on dense tabular structures, such as nutrition facts panels, and on multilingual ingredient lists with mixed fonts and orientations \cite{sporici2020}. Preprocessing steps such as grayscale conversion, contrast normalization, and denoising have been shown to reduce errors, yet challenges persist on real world packaging images that combine decorative elements with text \cite{guimaraes2023}.

Tesseract has been tested in food-label recognition tasks. Saputra et al. \cite{saputra2024} combined Tesseract with preprocessing and detection for nutrition-label extraction, reporting improved accuracy only after structured segmentation. These findings suggest that while Tesseract performs well on standardized layouts, it requires additional steps to maintain performance under uncontrolled packaging conditions.

\subsection{EasyOCR}
EasyOCR is an open-source OCR library implemented in Python and developed by JaidedAI, supporting over 80 languages \cite{jaided2020}. Its architecture combines convolutional layers for feature extraction and bidirectional recurrent layers for sequence modeling in a CRNN-based recognition network. When detection is enabled, EasyOCR uses the CRAFT detector, although recognition can run independently for pre-cropped or full images.

The main strength of EasyOCR lies in its multilingual support and lightweight deployment, which enables fast inference on GPU-based systems. These properties make it attractive for scenarios involving multi-language content, such as food packaging \cite{flores2024}. However, studies have reported that EasyOCR struggles with small fonts, cluttered backgrounds, and irregular layouts that combine multiple languages in dense panels \cite{guimaraes2023}. Even with multilingual capability, the absence of layout-specific optimizations limits its robustness on packaging images compared to controlled document text.

Prior evaluations show EasyOCR performing competitively on scene-text benchmarks but less consistently on structured domains. Flores et al. \cite{flores2024} confirmed that EasyOCR is more resilient than Tesseract under image distortions but still degrades when handling text mixed with graphical elements, a common condition in retail packaging.

\subsection{PaddleOCR}

PaddleOCR, developed by Baidu, is a modular OCR system designed for multilingual text and layout-aware recognition. It integrates DBNet-based detection with CNN-RNN recognition heads, while recent versions such as PP-OCRv5 incorporate lightweight models optimized for mobile and real-time inference. PaddleOCR also provides optional modules for document parsing, including PP-Structure, which can segment tables and key-value pairs \cite{du2020,cui2025}.

These capabilities make PaddleOCR theoretically suitable for packaging tasks that include structured layouts like nutrition tables. Rosyadi et al. \cite{rosyadi2024} demonstrated its practical use for ingredient-list extraction from smartphone images, reporting competitive accuracy but highlighting performance degradation in poor lighting and on reflective surfaces. While PaddleOCR supports angle classification for skew correction and multiple language packs, its effectiveness on unsegmented packaging images remains uncertain, as most prior work assumes either document-level or cropped-region inputs.

The inclusion of PaddleOCR in this study enables evaluation of a state-of-the-art modular system on full packaging images. This approach tests whether its advanced features for layout handling and multilingual recognition can compensate for the absence of explicit text-region detection in real world conditions.

\subsection{TrOCR}
TrOCR is a transformer-based OCR model introduced by Microsoft, combining a Vision Transformer (ViT) encoder with an autoregressive language decoder inspired by GPT-2. This architecture formulates OCR as a sequence-to-sequence task, eliminating handcrafted segmentation and allowing the model to learn contextual relationships across an image \cite{li2021}. TrOCR has demonstrated strong results on document and handwriting benchmarks, benefiting from large-scale pretraining and fine-tuning.

Despite its advantages, TrOCR lacks explicit layout modeling, which poses challenges for structured packaging images containing dense text blocks, small fonts, and multilingual content. Studies note that transformer-based OCR systems often exhibit output fragmentation when applied to cluttered or tabular layouts without preprocessing \cite{nagaonkar2025,baek2019}. Line-level segmentation or slicing is often necessary to prevent recognition collapse in complex layouts.

This evaluation includes TrOCR to assess whether its transformer-based design provides measurable gains on real world packaging or whether its reliance on full-image encoding limits accuracy without domain-specific tuning. Findings from prior work suggest significant trade offs: while transformers excel at contextual reasoning, they underperform in environments where layout and structure dominate recognition complexity \cite{nagaonkar2025}.

\section{Methodology}

\subsection{Dataset Description}

The dataset used in this study was collected in 2023 by the Department of Nutrition and Dietetics at the University of the Western Cape as part of the South African Nutrition Facts Panel (NFP) project. The collection contains multiple dated folders and subfolders. This paper uses one dated subfolder captured on 13~June~2023 (\textit{\texttt{20230613\_04\_03}}). Images were captured in retail environments using handheld mobile devices, introducing glare, reflections, and curved surfaces that can affect OCR accuracy. This subfolder contains 231 products and 1{,}628 images for timing and coverage, and accuracy is evaluated on a ground-truth subset of 113 images (60 products). Packaging in the selected subfolder includes multiple languages and mixed language strings, which increases recognition difficulty. Not all products include both an ingredient list and a nutrition facts panel; some have only one of these fields.

\begin{figure}[h]
\centering
\includegraphics[width=0.95\textwidth]{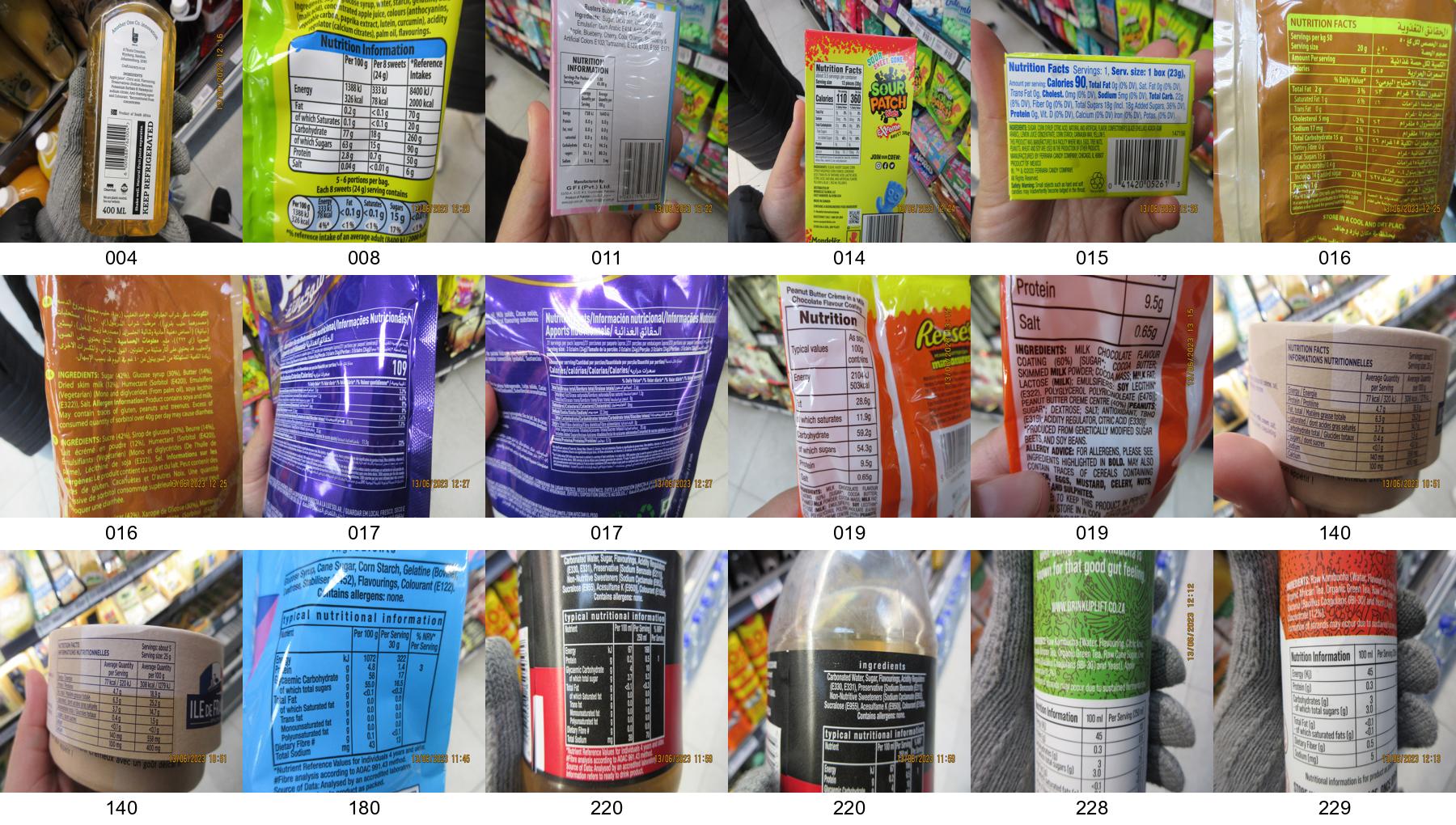}
\caption{Sample food packaging images from the SA NFP 2023 dataset.}
\label{fig:dataset_collage}
\end{figure}

Each product in this subfolder includes multiple views (2–23) showing different sides of the packaging, such as the front panel, ingredient list, and NFP. Images follow a structured naming format that encodes the capture date, session ID(XX), fieldworker ID(ZZZ), product ID(N), and image index as \textit{YYYYMMDD\_XX\_YY\_ZZZ (N).jpg}.  Figure~\ref{fig:dataset_collage} shows example images with product numbers, highlighting glare, curved surfaces, and multilingual ingredient lists that reflect the real world difficulty of this evaluation.

All 1,628 images were processed by the four OCR systems to measure execution time and coverage. For accuracy evaluation, a subset of 60 products (113 images) was manually transcribed to create ground truth for ingredient lists and NFP text. This design reflects practical deployment scenarios where processing speed is critical at scale, while detailed error metrics can only be computed for annotated data. The dataset differs significantly from available OCR benchmarks due to how packaging combines multilingual text, irregular layouts, and small fonts with decorative elements, making recognition more difficult than on documents or signage \cite{guimaraes2023,olejniczak2022}. This reflects broader trends in the South African packaged food supply, where visual clutter and on-package marketing often compromise the clarity of mandatory labeling \cite{khan2023}. This complexity makes this dataset well-suited for benchmarking OCR models under realistic deployment conditions.

\FloatBarrier

\subsection{OCR Models Used}
Four open-source OCR systems were evaluated in this study: Tesseract, EasyOCR, PaddleOCR, and TrOCR. These models represent distinct architectural paradigms—LSTM-based (Tesseract), CNN–RNN pipelines (EasyOCR and PaddleOCR), and transformer-based recognition (TrOCR). All systems were applied directly to full packaging images without prior region segmentation or keyword filtering to reflect realistic deployment conditions. This approach tests each model’s ability to handle common packaging challenges, such as multilingual text, small fonts, irregular layouts, and dense tabular structures.

\subsection{Implementation and Experimental Setup}

The experiments were conducted on a high-performance workstation running \textit{Windows 11 Enterprise (64-bit)} with an \textit{Intel 64-based processor} (24 cores, 32 threads) operating at 3.2\,GHz, 128\,GB RAM, and an \textit{NVIDIA GeForce RTX 3070 GPU} with 8\,GB of VRAM. All OCR systems were implemented in \textit{Python 3.11.9}. Key library versions were: \textit{pytesseract 0.3.13} for Tesseract, \textit{EasyOCR 1.7.2}, \textit{PaddleOCR 2.7.0.3}, and \textit{HuggingFace Transformers 4.53.3} for TrOCR. GPU acceleration was enabled for EasyOCR and TrOCR, while PaddleOCR ran on CPU due to backend compatibility limitations. Tesseract, by design, operates only on CPU \cite{smith2007,sporici2020}.

All four OCR systems were applied to full packaging images without region cropping or segmentation. Preprocessing for Tesseract included grayscale conversion, contrast enhancement, and non-local means denoising to enhance text clarity on cluttered layouts. EasyOCR was applied without additional custom steps beyond its internal resizing and normalization routines. PaddleOCR relied on its default PP-OCRv5 pipeline, which includes resizing, normalization, and angle classification for skew correction. TrOCR required RGB normalization because its Vision Transformer (ViT) encoder operates on three-channel inputs, and horizontal line-based slicing was applied before inference on dense areas such as nutrition facts tables to improve recognition accuracy compared to full-image inference \cite{li2021}.

\begin{figure}[h]
\centering
\includegraphics[width=0.9\textwidth]{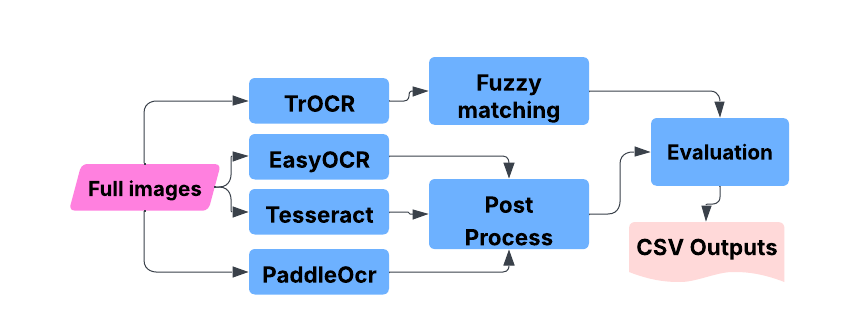}
\caption{OCR evaluation workflow from image input to Evaluations.}
\label{fig:ocr_workflow}
\end{figure}

Each system was executed with default models to avoid bias from fine-tuning. Tesseract employed its LSTM-based engine; EasyOCR used its CRNN recognizer with multilingual support; PaddleOCR ran with the lightweight PP-OCRv5 recognition model; and TrOCR utilized the pre-trained \texttt{microsoft/trocr-base-printed} configuration. Outputs were logged in CSV format with fields for \texttt{product\_id}, \texttt{image\_filename}, \texttt{raw\_ocr\_text}, and processing time. Figure~\ref{fig:ocr_workflow} provides an overview of the OCR evaluation pipeline, showing how full packaging images were processed through the four OCR systems, followed by normalization and metric-based evaluation. JSON outputs were also generated during initial testing for grouping checks but were excluded from final evaluation to maintain consistency across models.

\FloatBarrier

\subsection{Data Post-Processing and Normalization}

Raw OCR outputs were standardized to ensure comparability across models before evaluation. Predictions were initially stored in CSV format with fields such as \texttt{product\_id}, \texttt{image\_filename}, and \texttt{raw\_ocr\_text}, and then processed through a two-stage normalization pipeline.

For \textbf{Tesseract}, \textbf{EasyOCR}, and \textbf{PaddleOCR}, a keyword based classification strategy assigned text to either \textit{ingredients} or \textit{NFP}. When multiple candidates existed for the same product and text type, keyword density served as the tie-breaker to retain the most relevant block. This approach reflects common practices for handling structured text in OCR workflows where specific sections must be isolated for downstream tasks \cite{guimaraes2023,rosyadi2024}.

\textbf{TrOCR} required a different approach because its transformer encoder decoder architecture often produced fragmented predictions on dense packaging layouts. A fuzzy matching method using partial ratio scores was applied to identify text segments most associated with target keywords. A variant of the keyword density was also tested but produced incomplete matches, so fuzzy matching was preferred for this model.

After classification, the following normalization steps were applied across all systems:
\begin{itemize}
    \item Removal of special characters and redundant whitespace.
    \item Extraction of text following the first occurrence of relevant keywords (e.g., ``ingredients:'' or ``nutrition information'').
    \item Conversion to lowercase for uniform string comparison.
\end{itemize}

The normalized outputs were consolidated into standardized CSV files containing four fields: \texttt{product\_id}, \texttt{image\_filename}, \texttt{text\_type}, and \texttt{ocr\_text}. These files served as the input for all evaluation metrics described in Section~4.

\section{Evaluation Setup}

This section describes the metrics and procedures used to evaluate the OCR systems. The metrics used are widely used in OCR benchmarking and are standard for structured-text evaluations \cite{karatzas2013,neudecker2021}. Processing time per image was recorded for all models to assess efficiency.

\subsection{Character Error Rate (CER)}
CER quantifies character-level accuracy by computing the Levenshtein distance between the predicted and reference text, normalized by the length of the ground truth \cite{karatzas2013,neudecker2021}. It penalizes insertions, deletions, and substitutions equally. CER is defined as:
\[
\text{CER} = \frac{S + D + I}{N},
\]
where $S$ represents substitutions, $D$ deletions, $I$ insertions, and $N$ the number of characters in the reference string.

\subsection{Word Error Rate (WER)}
WER applies the same principle as CER but operates at the word level, reflecting the preservation of semantic units in OCR output. It is particularly relevant for ingredient lists and nutrition facts panels \cite{karatzas2013}. WER is defined as:
\[
\text{WER} = \frac{S + D + I}{N_w},
\]
where $N_w$ denotes the number of words in the reference text.

\subsection{BLEU (Bilingual Evaluation Understudy)}
BLEU measures n-gram overlap between predicted and reference text. Although originally introduced for machine translation, BLEU is widely applied in OCR research for evaluating structural fidelity \cite{papineni2002}. The metric is calculated as:
\[
\text{BLEU} = BP \cdot \exp \left( \sum_{n=1}^k w_n \log p_n \right),
\]
where $p_n$ is the modified n-gram precision, $w_n$ the weight for each n-gram size, and $BP$ the brevity penalty.

\subsection{ROUGE-L (Recall-Oriented Understudy for Gisting Evaluation)}
ROUGE-L evaluates text similarity using the longest common subsequence (LCS), combining precision and recall to account for sequence order. This property makes ROUGE-L effective for structured text such as ingredient lists \cite{lin2004}. It is defined as:
\[
\text{ROUGE-L}_{F_1} = \frac{(1+\beta^2) \cdot R \cdot P}{R + \beta^2 P},
\]
where $R$ denotes recall, $P$ precision, and $\beta$ is typically set to 1 for equal weighting.

\subsection{F1 Score}
The F1 score, calculated at the word level, represents the harmonic mean of precision and recall, emphasizing exact word matches. It complements BLEU and ROUGE by focusing on strict token overlap, which is important for structured fields such as nutrition panels \cite{karatzas2013}. The formula is:
\[
\text{F1} = \frac{2 \cdot \text{Precision} \cdot \text{Recall}}{\text{Precision} + \text{Recall}}.
\]

\subsection{Implementation Details}
All metrics were implemented in Python 3.11.9 using the following libraries: \texttt{python-Levenshtein} for CER and WER, \texttt{nltk.translate.bleu\_score} for BLEU (with smoothing method 4), \texttt{rouge\_score} for ROUGE-L, and a custom implementation for F1 at the word level. Normalization steps prior to evaluation included keyword-based classification for Tesseract, EasyOCR, and PaddleOCR, fuzzy matching for TrOCR, removal of punctuation, case folding, and trimming redundant characters (see Section~3.4). Keyword based sectioning on full images can include nearby lines (for example allergen or storage text); this increases insertions and can raise CER on short strings.

Accuracy metrics were computed using 113 manually transcribed images from 60 products, representing the ground truth subset. Processing time per image is measured across all 1{,}628 images to reflect realistic workload conditions. Accuracy metrics are computed per image on the same ground truth images (n=113), so models are compared on identical samples; Coverage is reported at two levels: product level counts a product as covered if at least one present target field (ingredient list or NFP) yields non-empty OCR after sectioning. Field level coverage is computed separately for ingredient lists and for nutrition facts panels, using as denominator only products that contain the respective field. Evaluation treats text uniformly regardless of language or mixing, and character and word error are computed directly against the ground truth without language specific rules. Despite the availability of newer neural-based similarity measures, BLEU and ROUGE-L remain widely adopted in OCR evaluation due to their interpretability and established benchmarks, building on their original definitions \cite{papineni2002,lin2004}.

\section{Results and Discussion}

This section presents the evaluation results for the four OCR systems: EasyOCR, Tesseract, PaddleOCR, and TrOCR. Performance was evaluated using character-level metrics (CER and WER), semantic metrics (BLEU, ROUGE-L, and F1), coverage, and execution time. The findings are interpreted in relation to dataset characteristics and model architectures to assess their suitability for food packaging text extraction in real world conditions.

\subsection{Overall Performance Summary}

The evaluation used 60 products with 113 images, covering both ingredient lists and NFP. These samples were selected from the full set of 231 products to maintain diversity while keeping the evaluation manageable. Table~\ref{tab:semantic_performance} presents the mean BLEU, ROUGE-L, F1 scores, and coverage for each OCR model.

Tesseract achieved the highest BLEU, ROUGE-L, and F1 scores, showing the strongest semantic accuracy among all systems. PaddleOCR achieved the highest coverage among the convolution-based models and delivered competitive ROUGE-L performance. EasyOCR provided balanced results across all metrics, while TrOCR produced the lowest semantic scores despite full coverage, highlighting its limitations when handling dense layouts and multilingual content.

\begin{table}[h]
\centering
\caption{Overall semantic performance of OCR models. Best values in each column are in \textbf{bold}.}
\label{tab:semantic_performance}
\begin{tabular}{|l|c|c|c|c|}
\hline
\textbf{Model} & \textbf{Coverage (\%)} & \textbf{BLEU} & \textbf{ROUGE-L} & \textbf{F1} \\
\hline
EasyOCR & 91.53 & 0.153 & 0.314 & 0.265 \\
\textbf{Tesseract} & 79.66 & \textbf{0.245} & \textbf{0.391} & \textbf{0.345} \\
PaddleOCR & 98.31 & 0.163 & 0.361 & 0.248 \\
TrOCR & \textbf{100.00} & 0.010 & 0.026 & 0.017 \\
\hline
\end{tabular}
\end{table}
 
\subsection{CER and WER Analysis}

\begin{figure}[h]
\centering
\includegraphics[width=0.9\textwidth]{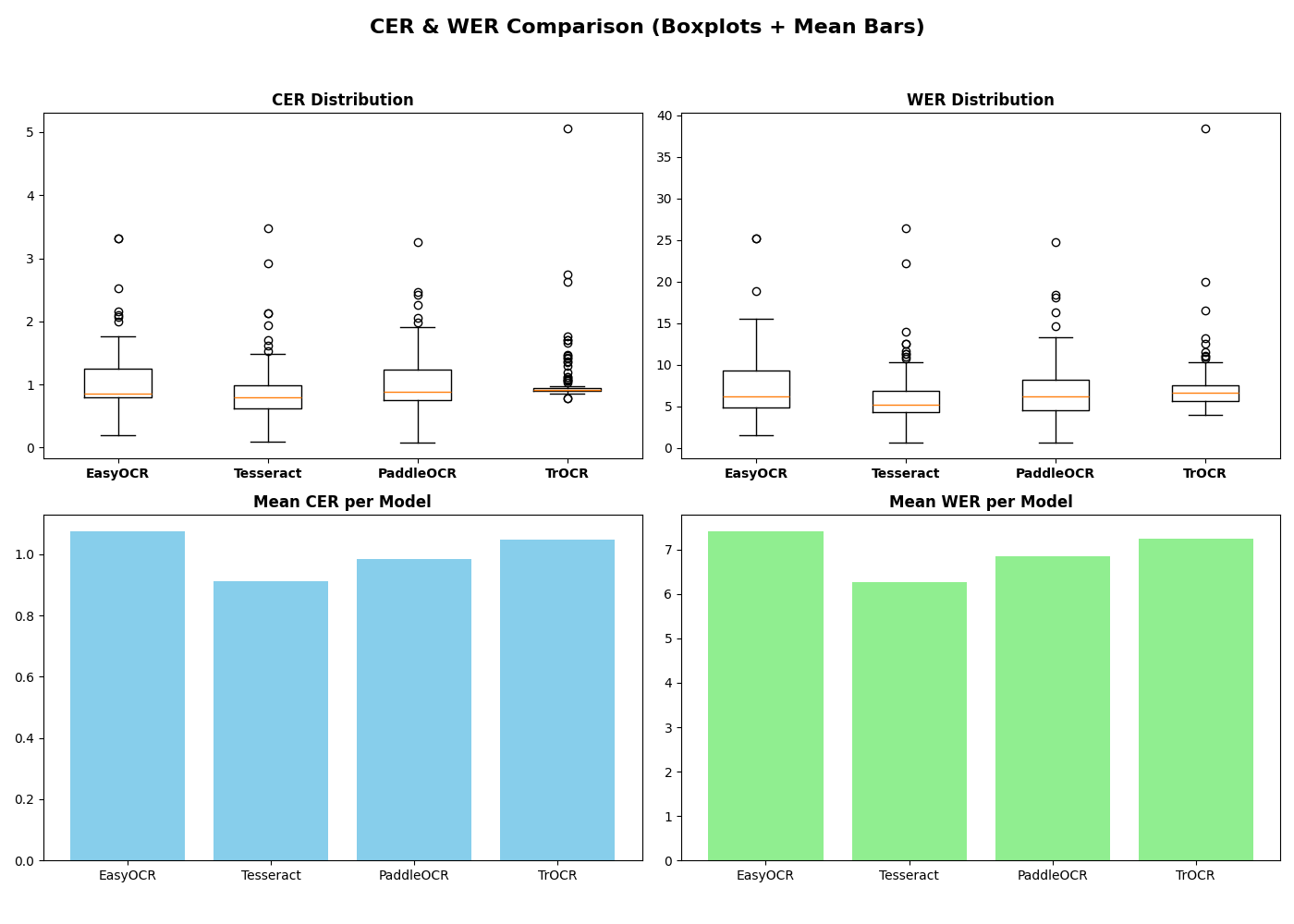}
\caption{CER and WER with box plots for distributions and bars for mean values.}
\label{fig:cer_wer_mixed}
\end{figure}

Character Error Rate (CER) and Word Error Rate (WER) evaluate recognition accuracy at the character and word levels. The per image CER and WER distributions for all models are shown in Figure~\ref{fig:cer_wer_mixed}. Table~\ref{tab:cer_wer} reports the corresponding mean values over images for quick reference. Higher rates of insertions and substitutions are observed on mixed language strings, which increases CER and WER on short segments. Character error rate follows the normalized edit distance definition used in OCR (see Section 4) \cite{neudecker2021,baek2019}. Values above 1.0 can occur on short references when insertions dominate or when the extracted span includes adjacent text.




Across the per image distributions in Figure~\ref{fig:cer_wer_mixed}, Tesseract shows lower central tendency and a tighter spread on both metrics on this dataset. EasyOCR and PaddleOCR have intermediate error rates with occasional outliers, and they process a larger share of products in the coverage analysis. TrOCR exhibits the widest error range, with WER values up to 38.4, which aligns with sensitivity to complex layouts. Results indicate a trade off between recognition accuracy and coverage, so the preferred model depends on the primary requirement. Differences in central tendency are small for some pairs and the distributions overlap, so practical choice should consider coverage and throughput alongside accuracy.

\begin{table}[h]
\centering
\caption{Per-image CER and WER (mean $\pm$ SD) values over images per model.}
\label{tab:cer_wer}
\begin{tabular}{|l|r|r|}
\hline
\textbf{Model} & \textbf{CER mean $\pm$ SD} & \textbf{WER mean $\pm$ SD} \\
\hline
EasyOCR   & 1.075 $\pm$ 0.569 & 7.408 $\pm$ 4.253 \\
PaddleOCR & 0.985 $\pm$ 0.539 & 6.857 $\pm$ 4.004 \\
Tesseract & 0.912 $\pm$ 0.584 & 6.262 $\pm$ 4.247 \\
TrOCR     & 1.049 $\pm$ 0.484 & 7.243 $\pm$ 3.759 \\
\hline
\end{tabular}
\end{table}

\FloatBarrier


\subsection{BLEU, ROUGE-L, and F1 Analysis}

\begin{figure}[h]
\centering
\includegraphics[width=0.9\textwidth]{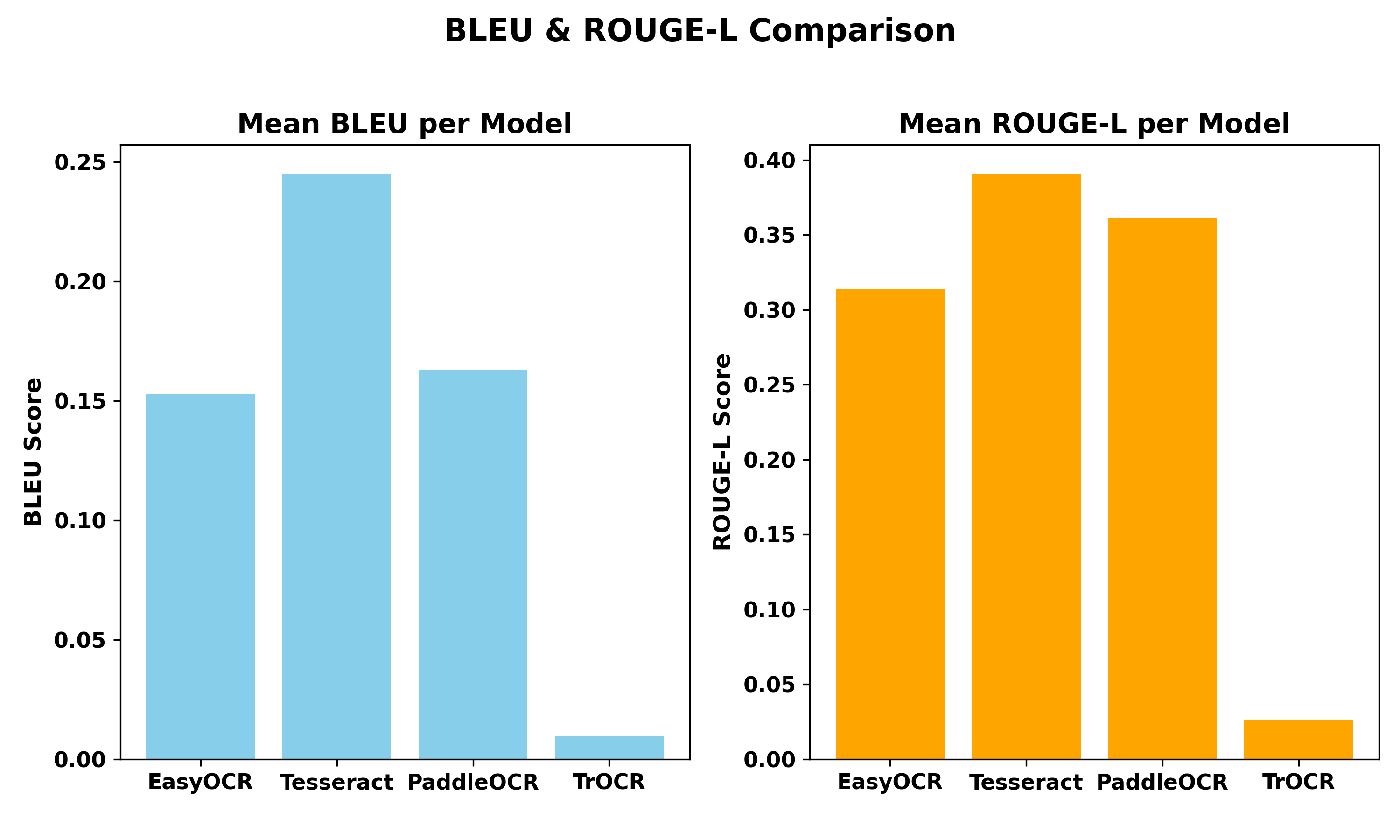}
\caption{BLEU and ROUGE-L comparison across OCR models.}
\label{fig:bleu_rouge}
\end{figure}

\begin{figure}[h]
\centering
\includegraphics[width=0.9\textwidth]{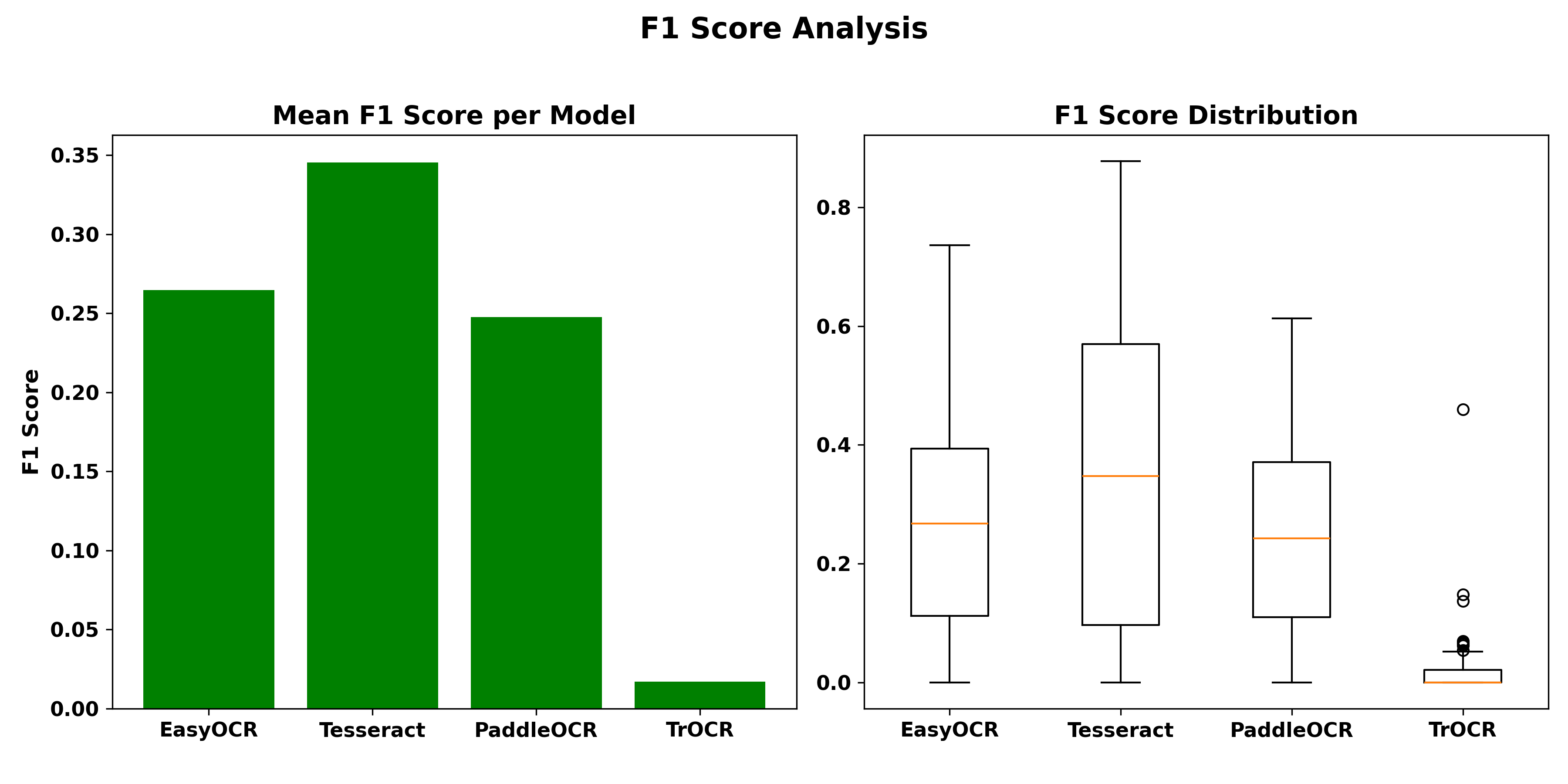}
\caption{F1 score distribution across OCR models.}
\label{fig:f1_scores}
\end{figure}

BLEU and ROUGE-L measure n-gram and sequence-based similarity, while F1 evaluates precision and recall at the token level. Figure~\ref{fig:bleu_rouge} compares BLEU and ROUGE-L scores, and Figure~\ref{fig:f1_scores} shows F1 performance. BLEU was designed for corpus-level evaluation \cite{papineni2002}, and ROUGE-L measures n-gram overlap \cite{lin2004}; on short packaging strings small ordering and token differences remove many n-grams, so low scores are expected and comparisons emphasise CER and WER.

Tesseract achieved the highest BLEU and ROUGE-L scores (0.245 and 0.391), confirming its advantage in preserving lexical and structural accuracy. EasyOCR and PaddleOCR delivered moderate scores, while TrOCR recorded very low BLEU (0.010) and ROUGE-L (0.026) values, indicating limited capability for structured text on packaging. These outcomes suggest that convolution-based models are currently more reliable than transformer-based architectures when applied to complex layouts without fine-tuning.

Although new semantic similarity metrics have emerged, BLEU and ROUGE remain standard for OCR text fidelity evaluation because of their interpretability and benchmarking history, as established by Papineni et al. \cite{papineni2002} and Lin \cite{lin2004}. BLEU and ROUGE L are lower on short packaging strings, so accuracy comparisons emphasize CER and WER.

\FloatBarrier

\FloatBarrier

\subsection{Coverage Analysis}
Coverage refers to the percentage of products for which an OCR model returned non empty text for at least one target field present on that product (ingredient list or NFP). Products that lack one of the fields are evaluated on the fields that exist. TrOCR achieved full coverage at 100\%, followed by PaddleOCR at 98.31\%, EasyOCR at 91.53\%, and Tesseract at 79.66\%. The complete coverage recorded by TrOCR indicates that the model consistently returned text outputs, although these outputs often lacked semantic accuracy. In contrast, the lower coverage observed for Tesseract is associated with its sensitivity to image noise and stylized fonts, which occasionally resulted in empty outputs.

\subsection{Execution Time Comparison}

Average processing times for each model are shown in Table~\ref{tab:execution_time}. Tesseract was the fastest system, with an average of 0.58 seconds per image, making it well-suited for large-scale deployments on CPU-based environments. EasyOCR followed with an average time of 0.81 seconds per image, benefiting from GPU acceleration on the NVIDIA RTX\,3070. PaddleOCR was the slowest, averaging 6.24 seconds per image due to its execution in CPU-only mode, which was required because of PaddlePaddle GPU compatibility limitations. TrOCR averaged 2.20 seconds per image despite running on GPU, reflecting the higher computational cost of its Transformer-based architecture.

\begin{table}[h]
\centering
\caption{Average execution time per image and hardware utilization.}
\label{tab:execution_time}
\begin{tabular}{|l|c|c|l|}
\hline
\textbf{Model} & \textbf{Total Time (s)} & \textbf{Avg Time/Image (s)} & \textbf{Hardware Used} \\
\hline
EasyOCR & 1,322.10 & 0.81 & GPU (RTX\,3070) \\
\textbf{Tesseract} & \textbf{949.53} & \textbf{0.58} & CPU Only \\
PaddleOCR & 10,161.16 & 6.24 & CPU Only \\
TrOCR & 3,573.58 & 2.20 & GPU (RTX\,3070) \\
\hline
\end{tabular}
\end{table}

\FloatBarrier

\subsection{Discussion}

The evaluation shows that Tesseract achieved the highest overall accuracy across character-level metrics (CER and WER) and semantic measures (BLEU, ROUGE-L, and F1). Its LSTM-based recognition pipeline appears well-suited for structured text such as ingredient lists and nutrition panels. This outcome aligns with the findings of Smith \cite{smith2007}, who reported that Tesseract remains effective on printed text when combined with basic preprocessing.

EasyOCR and PaddleOCR produced competitive results. PaddleOCR obtained the highest coverage but had the longest processing times because it ran exclusively on CPU during these experiments. Although the system supports GPU acceleration, compatibility issues with CUDA prevented its use in this setup, which is a known limitation documented in practice \cite{rosyadi2024}. This constraint, combined with its multi-stage architecture, contributed to slower execution. EasyOCR offered a more balanced trade off between speed and accuracy, particularly in GPU-enabled environments, making it suitable for multilingual packaging scenarios \cite{jaided2020}.

TrOCR recorded the weakest results despite GPU acceleration and a transformer-based architecture. Similar limitations were noted by Li et al. \cite{li2021}, who emphasized that transformer-based OCR models often require domain-specific fine-tuning to manage irregular layouts and dense structures effectively. Tesseract's advantage over TrOCR can be attributed to its optimization for printed text and its ability to process structured layouts with minimal adjustments \cite{smith2007,sporici2020}. TrOCR, by contrast, was primarily developed for scanned documents and handwriting rather than multi-column packaging layouts \cite{li2021}. These design characteristics likely explain the fragmented outputs and low semantic accuracy observed in this study.

The results have practical implications for real world deployment. High OCR accuracy is essential for regulatory compliance, as errors in ingredient or nutritional information may lead to health and legal risks \cite{guimaraes2023}. Recent evaluations of label compliance in South Africa have shown that many food packages do not present nutrition information clearly, even to human readers \cite{abdoolkarim2025}. This is especially important in products marketed to children, where visual distractions and misleading branding can obscure key information \cite{khan2023}. Reliable text extraction also enables the development of digital nutrition databases, mobile health applications, and large-scale dietary research \cite{seitaj2024}. Based on these findings, Tesseract is most suitable for CPU-based environments that require high accuracy, EasyOCR is recommended for GPU-enabled scenarios with multilingual requirements, and PaddleOCR is appropriate where advanced layout handling is needed and GPU resources are available. TrOCR may only be suitable in specialized use cases that allow model fine-tuning and benefit from transformer interpretability.

\section{Conclusion}

This study evaluated Tesseract, EasyOCR, PaddleOCR, and TrOCR on food packaging images from South African retail environments, focusing on ingredient lists and nutrition facts panels. Performance was measured using character and word error, BLEU, ROUGE-L, F1, coverage, and execution time. On this dataset, Tesseract achieved the lowest mean CER and WER, while EasyOCR provided balanced performance with strong multilingual capability. PaddleOCR delivered the highest coverage but required significantly longer processing times because it was limited to CPU execution in this setup. TrOCR, although GPU accelerated, recorded the lowest accuracy, highlighting the challenges of applying transformer-based models to complex packaging layouts without domain-specific tuning. On this dataset, Tesseract shows the lowest mean CER and WER on the ground truth subset. EasyOCR and PaddleOCR cover a larger share of products, while TrOCR underperforms on accuracy. These results indicate a trade off between recognition accuracy and coverage, the preferred system depends on the primary requirement of the application.

The results provide a reference point for OCR performance on packaging data and can assist in selecting suitable systems for applications such as regulatory compliance, nutrition databases, and mobile health platforms. Future work should focus on fine-tuning OCR models for packaging text, integrating region-aware detection methods, and improving post-processing to better handle multilingual content and complex layouts.

\bibliographystyle{unsrt}  
\bibliographystyle{splncs04}

\end{document}